\documentclass{article}

\PassOptionsToPackage{numbers,sort}{natbib}
\PassOptionsToPackage{table,x11names}{xcolor}
\usepackage[preprint]{neurips_2026}

\usepackage[utf8]{inputenc}
\usepackage[T1]{fontenc}
\usepackage{hyperref}
\usepackage{booktabs}
\usepackage{amsfonts}
\usepackage{amsmath}
\usepackage{amssymb}
\usepackage{nicefrac}
\usepackage{microtype}
\usepackage{graphicx}
\usepackage{subcaption}
\usepackage{pifont}

\usepackage{enumitem}
\usepackage{multirow}
\usepackage{caption}
\usepackage[table,xcdraw]{xcolor}
\usepackage[normalem]{ulem}
\useunder{\uline}{\ul}{}
\usepackage{wrapfig}

\usepackage{wrapfig}
\usepackage{liauto_style}
\liautologo{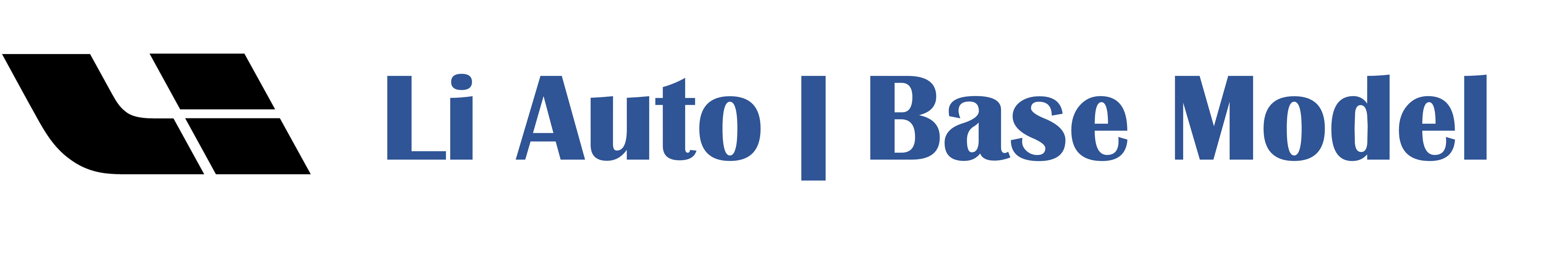}

\newcommand{\blockcomment}[1]{}

\title{Dual-Pathway Geometry-Aware MLLM for Spatial Intelligence}

\author{%
  Yufei Zheng\textsuperscript{1,2}\quad
  Xuhan Zhu\textsuperscript{2}\quad
  Zide Liu\textsuperscript{2}\quad
  Chunpeng Zhou\textsuperscript{2}\quad
  Chenfeng Wang\textsuperscript{1,2}\quad
  Yongchao Xu\textsuperscript{1}\\
  Yunnan Wang\textsuperscript{3}\quad
  Jiawei Liu\textsuperscript{1,$\dagger$}\quad
  Pengfei Yu\textsuperscript{2,$\dagger$}\quad
  Wei Zhai\textsuperscript{1,$\dagger$}\quad
  Yang Cao\textsuperscript{1}\\
  Zheng-Jun Zha\textsuperscript{1} \\
  \\
  \mdseries\textsuperscript{1}University of Science and Technology of China\quad
  \textsuperscript{2}Li Auto Inc.\quad
  \mdseries\textsuperscript{3}Shanghai Jiao Tong University \\
}

\begin{document}
\maketitle
{\renewcommand{\thefootnote}{$\dagger$}\footnotetext{Corresponding author.}}

\begin{liautoabstract}
Spatial understanding of the physical world from 2D visual inputs hinges on two complementary forms of geometric knowledge: holistic 3D structural perception and fine-grained metric scale estimation. Existing multimodal large language models (MLLMs) typically address only one facet, ingesting either depth maps or point clouds as additional model inputs, which incurs substantial computational overhead and inherits the generalization limitations of upstream prediction models. We propose GAMSI, a dual-pathway Geometry-Aware MLLM for Spatial Intelligence that takes only RGB images as input while internalizing both forms of geometric prior within a unified autoregressive backbone. Specifically, we introduce Metric-Structure Decoupled Queries (MSDQ) which employ two groups of learnable queries to respectively extract dense metric signals and sparse structural cues from the shared visual context, with a task-decoupled attention mask further preventing the two pathways from contaminating each other. Building on this, an Expert-Guided Visual Grounding (EVG) module projects the aggregated cues back to frame-level visual features and aligns them with vision foundation models, which serve purely as training-time supervision, rather than as model inputs. We further build a multi-task spatial instruction-tuning dataset (MTS) comprising 152{,}776 samples spanning 13 task types and three visual modalities, consolidated from six public datasets. Trained with a two-stage curriculum, GAMSI achieves state-of-the-art performance on seven spatial intelligence benchmarks.

\vspace{3mm}
    {\color{liautoblue!30}\rule{\linewidth}{0.5pt}} 
    \vspace{2mm}

    \small 
    \renewcommand{\arraystretch}{1.3} 
\begin{tabular}{@{} l l @{}}
        {\color{liautoblue}\faCalendar*} & \textbf{Date:} May 25, 2026 \\
        {\color{liautoblue}\faEnvelope} & \textbf{Correspondence:} {jwliu6@ustc.edu.cn, yupengfei1@lixiang.com, wzhai056@ustc.edu.cn} \\
        {\color{liautoblue}\faGithub} & \textbf{Project Page:} \url{https://github.com/zzzyf01/GAMSI} \\
\end{tabular}

\end{liautoabstract}

\section{Introduction}

Spatial understanding \cite{yang2025thinking,wu2025spatial,li2025spatialladder,chen2024spatialvlm} refers to the ability to perceive 3D structure, infer metric scale, and reason about geometric relationships from 2D observations. It underpins a wide range of downstream applications, including embodied manipulation \cite{song2026maniplvm,kim2024openvla}, autonomous navigation \cite{nahavandi2025comprehensive}, and AR/VR interaction \cite{huang2024virtualnexus}. In these settings, a system must not only recognize objects but also quantify their distances, sizes, and spatial arrangements, often from limited visual input.
Recently, Multimodal Large Language Models (MLLMs) \cite{yang2025visual,liu2024improved,zhu2023minigpt,alayrac2022flamingo,bai2025qwen3,zhu2025internvl3} have emerged as the default interface for such tasks, unifying perception, language, and reasoning within a single autoregressive backbone queried via natural language. Whether MLLMs genuinely possess such spatial cognition, rather than merely producing vague, ungrounded spatial descriptions, has become a central research question in the field.

\begin{figure}[t]
  \centering
  \includegraphics[width=1.0\textwidth]{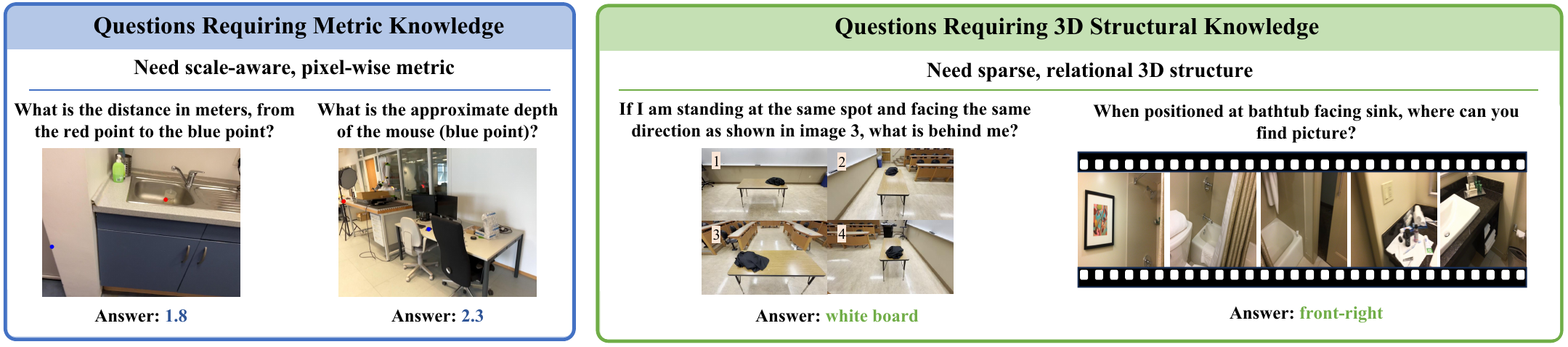}
  \caption{Two complementary categories of spatial questions. Metric questions require scale-aware, pixel-wise cues, whereas 3D structural questions require sparse, relational cues.}
  \label{fig1}
\end{figure}

Despite the encouraging progress made by existing studies, endowing MLLMs with reliable spatial cognition remains fundamentally challenging. A closer examination of real-world spatial questions reveals that they essentially fall into two complementary categories \cite{wu2025spatial,cai2025spatialbot}, each demanding a distinct form of geometric knowledge that purely 2D semantic features alone are insufficient to provide, as illustrated in Fig.~\ref{fig1}.
(i) A large class of spatial questions concerns absolute physical quantities, such as asking the distance in meters between two points or the depth of a specific object. Answering them relies on scale-aware, pixel-wise metric cues, as produced by monocular depth estimators \cite{yang2024depth,yang2024depthv2}, where every pixel is anchored to a concrete physical unit. Without such a dense, unit-bearing signal, MLLMs can offer at best qualitative guesses, since RGB features alone suffer from inherent scale ambiguity.
(ii) An equally important class involves relational, viewpoint-dependent geometric information, such as inferring what lies behind the camera or locating an object from a specified viewpoint. Answering them depends on sparse, relational 3D structural cues, as recovered by multi-view 3D reconstruction models such as VGGT \cite{wang2025vggt}, which encode inter-frame correspondences, camera motion, and global scene layout in a point-cloud-like feature space. Such structural information differs fundamentally from pixel-wise depth: it is relational rather than dense, and viewpoint-aware rather than unit-bound.

However, existing approaches typically address only one facet of this picture. The first family of methods \cite{cai2025spatialbot,liu2025ssr,daxberger2025mm} feeds predicted depth maps into the model as an additional input modality; while such methods improve the perception of distance and size, they offer little benefit for relational or viewpoint-dependent questions. The second family \cite{wu2025spatial,hong20233d} instead exploits 3D structural priors, such as explicit point clouds or features extracted from 3D reconstruction foundation models; these methods strengthen layout and viewpoint understanding, yet lack the absolute scale information required for metric tasks. To date, very few works have sought to jointly leverage these two complementary sources of geometric knowledge. Moreover, the vast majority of existing methods \cite{cai2025spatialbot,wu2025spatial} rely on auxiliary prior models to generate additional information such as depth maps or 3D point clouds, and feed them into the model as part of its input. This not only amplifies dependence on the generalization capability of external models, but also substantially increases the inference-time overhead.

To address the above limitations, we propose GAMSI, a dual-pathway Geometry-Aware MLLM framework for Spatial Intelligence. Unlike prior approaches, GAMSI operates on raw RGB images alone at inference, requiring no depth maps, point clouds, or camera parameters, and thus removes any inference-time dependence on auxiliary inputs or external models. Its core idea is to route two heterogeneous types of geometric knowledge through two decoupled pathways and internalize these priors into the MLLM's own semantic space during training, enabling the model to spontaneously invoke the corresponding geometric cognition from images alone at inference. To this end, we introduce two sets of learnable query embeddings dedicated to metric depth and 3D structure, respectively. Since a naive causal mask would allow the two query groups to leak into each other via shared self-attention, our Metric-Structure Decoupled Queries (MSDQ) adopt a task-decoupled attention mask under which each group attends only to the image tokens while remaining mutually invisible, cleanly disentangling the dense metric signal from the sparse structural cue at the representation level. Furthermore, since the queried information resides in the LLM's abstract semantic space and is inherently misaligned with the geometric feature space required for precise spatial perception, we further introduce an Expert-guided Visual Grounding (EVG) module, in which the visual queries project the aggregated depth and 3D cues into a frame-level feature space supervised by pretrained vision foundation models during training only, leaving inference entirely free of any external model.

To progressively inject and activate the dual-path geometric priors of GAMSI, we adopt a two-stage training paradigm that advances from low-level perceptual alignment to high-level task-oriented intelligence. In the first stage, we leverage the SenseNova-SI-800K dataset~\cite{cai2025scaling} to align the MSDQ and EVG modules via metric-depth and 3D-structure supervision. However, SenseNova-SI is perception-centric by design: its samples are heavily concentrated on a handful of foundational tasks such as general visual understanding, metric measurement, and perspective-taking, while higher-order categories such as spatial imagination, object counting, and appearance-order recall are extremely scarce, leaving its instruction-format coverage too narrow on its own. To bridge this gap, in the second stage we construct MTS (Multi-Task Spatial), an instruction-tuning dataset consolidated from six complementary public sources~\cite{daxberger2025mm, yin2025spatial, zhang2025flatland, liu2025can, li2025spatialladder, yang2025cambrian} to replenish the high-order task samples missing from SenseNova-SI. MTS contains 152{,}776 instances across 13 curated task types, organized into metric and structural spatial perception and spanning both simple and complex tasks to jointly supervise the two branches of GAMSI. It is further balanced across single-image, multi-image, and video modalities to prevent overfitting to any particular input form, and will be publicly released to support future research on general-purpose spatial intelligence.

Our contributions are summarized as follows: (1) We propose GAMSI, a dual-pathway geometry-aware MLLM that internalizes metric-depth and 3D structural priors into a single autoregressive backbone, enabling reliable spatial cognition from RGB images alone. (2) We introduce two sets of learnable queries with a task-decoupled attention mask that disentangles metric and structural signals, preventing cross-contamination under shared self-attention. (3) We design an EVG module in which visual queries project the depth and 3D cues aggregated within the LLM back to the visual space, supervised by vision foundation models. (4) We build MTS, a multi-task spatial intelligence dataset of 152{,}776 samples covering 13 tasks and three modalities. Extensive experiments show that GAMSI achieves state-of-the-art performance across diverse spatial intelligence benchmarks.

\section{Related Work}
\textbf{Visual Spatial Intelligence.}
Enhancing the spatial understanding capability of MLLMs has attracted increasing attention. Extensive benchmark studies \cite{yang2025thinking,li2025spatialladder,tong2024cambrian} reveal that, despite notable progress on general visual tasks, existing models still exhibit clear deficiencies in spatial understanding. Prior efforts can be broadly grouped into two categories. The first feeds predicted depth maps as an auxiliary input \cite{cai2025spatialbot,liu2025ssr,daxberger2025mm}, which improves the perception of distance and size but offers little benefit for relational or viewpoint-dependent questions. The second exploits 3D structural priors from reconstruction foundation models, as in 3DSR \cite{huang2025mllms}, 3DThinker \cite{chen2025think}, and Spatial-MLLM \cite{wu2025spatial}, strengthening layout and viewpoint understanding but lacking the absolute scale needed for metric tasks. Very few works jointly leverage these two complementary cues, leaving the modeling of geometric layout and depth scale underexplored. Moreover, most existing methods \cite{cai2025spatialbot,wu2025spatial} rely on auxiliary prior models to produce depth maps or point clouds as inputs, which amplifies dependence on external models and increases inference-time overhead. In contrast, we propose a spatial intelligence architecture that takes only images and text as input, requiring no auxiliary 3D data, and adaptively extracts both 3D structural features and metric depth features from visual inputs to jointly support spatial perception. With this design, our model achieves a unified understanding of geometric layout and depth scale in a purely RGB-based, annotation-efficient paradigm.

\textbf{Vision Foundation Model.}
Vision foundation models (VFMs) have become a cornerstone of modern visual understanding, producing transferable representations that adapt to a wide range of downstream tasks \cite{daxberger2025mm,wang2025vision}. On the semantic side, contrastive vision-language models such as CLIP~\citep{radford2021learning} and ALIGN~\citep{jia2021scaling} align images with natural language at scale and endow visual encoders with open-vocabulary priors; self-supervised models such as DINO~\citep{caron2021emerging} and DINOv2~\citep{oquab2023dinov2} learn dense features with strong semantic information and part-level structure; promptable segmentation models such as SAM~\citep{kirillov2023segment} and SAM~2~\citep{ravi2024sam} further provide class-agnostic mask priors that can be flexibly plugged into perception pipelines. In parallel, a growing body of geometry-oriented foundation models extends this paradigm to 3D: monocular depth predictors such as ZoeDepth~\citep{bhat2023zoedepth}, and Depth~Anything~\citep{yang2024depth,yang2024depthv2} achieve zero-shot depth estimation by training on massive heterogeneous data, while feed-forward reconstruction models such as MASt3R~\citep{leroy2024grounding}, and VGGT~\citep{wang2025vggt} unify pose estimation, dense matching, and point cloud regression within a single network, turning multi-view geometry into a learnable prior. Building on this line of research, our work integrates the capabilities of these VFMs in monocular depth estimation and 3D point cloud reconstruction into a unified large model, which directly extracts holistic spatial layout and metric depth information from raw images and uses them as structured priors to support spatial intelligence and spatial understanding tasks.

\begin{figure}[t]
  \centering
  \includegraphics[width=1.0\textwidth]{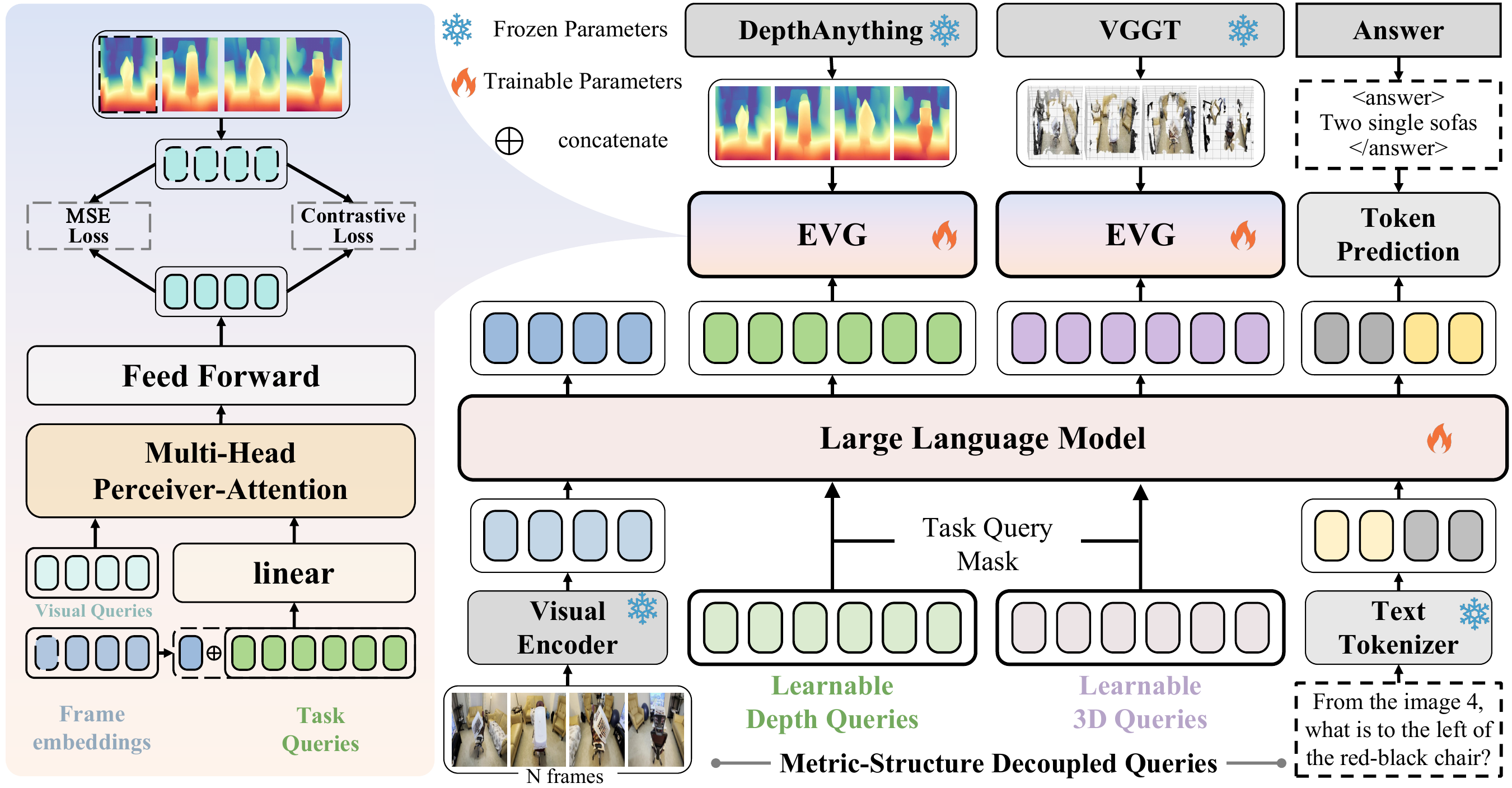}
  \caption{The overall architecture of the proposed GAMSI. It consists of two key modules: (1) Metric-Structure Decoupled Queries (MSDQ): routes two sets of learnable queries into independent pathways via a task-decoupled attention mask, separately encoding metric-depth and 3D structural cues. (2) Expert-Guided Visual Grounding (EVG): anchors each pathway to the feature space of a complementary visual expert (Depth Anything and VGGT), injecting task-specific priors for reliable spatial understanding.}
  \label{fig2}
\end{figure}

\section{Methodology}
\label{method}
As illustrated in Figure~\ref{fig2}, GAMSI jointly learns 3D structural and metric-depth representations from image–text inputs through two tightly coupled designs. The first is Metric-Structure Decoupled Queries (MSDQ), a dual-pathway query mechanism that employs task-decoupled attention to independently encode geometry-level structural cues and scale-aware metric-depth cues, while a shared self-attention backbone preserves cross-pathway interaction. The second is Expert-Guided Visual Grounding (EVG), which anchors each pathway's queries in the feature spaces of complementary visual experts, namely a 3D reconstruction model for structural grounding and a metric depth estimator for scale grounding, thereby injecting pathway-specific priors. Built upon a general MLLM backbone, GAMSI is trained in two stages: spatial perception alignment, followed by multi-task spatial understanding fine-tuning.

\subsection{Metric-Structure Decoupled Queries}

\begin{wrapfigure}{r}{0.51\textwidth}
  \centering
  \vspace{-10pt}
  \includegraphics[width=0.51\textwidth]{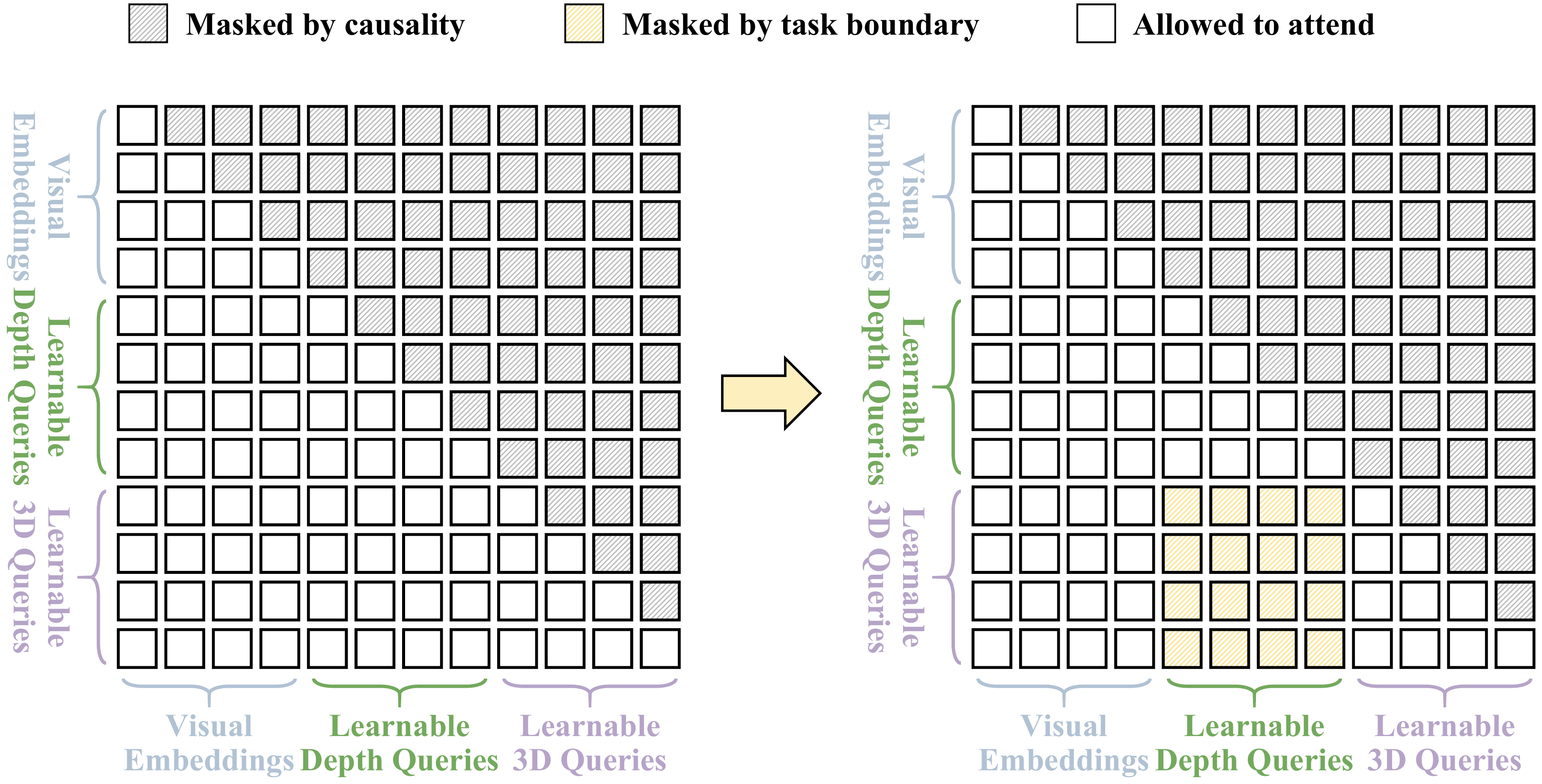}
  \caption{
  Comparison between the original causal mask (left) and our task-decoupled mask (right), which additionally blocks cross-task attention between Depth and 3D queries (yellow).
  }
  \label{fig3}
  \vspace{-10pt}
\end{wrapfigure}

To extract two complementary forms of spatial representations directly from image–text inputs, we introduce two sets of task-specific learnable query embeddings into the LLM input sequence.
Formally, given a scene \(\mathcal{S} = \{I^i\}_{i=1}^N\) consisting of \(N\) frames with \(I^i \in \mathbb{R}^{H \times W \times 3}\), a textual question \(\mathcal{Q}\), and its corresponding answer \(\mathcal{A}\), the vision encoder produces per-frame patch embeddings \(F_v = \{f_v^i\}_{i=1}^N\), where \(f_v^i \in \mathbb{R}^{P \times C}\).
On the text side, \(\mathcal{Q}\) is first tokenized into a sequence of token ids and then mapped through the token embedding layer to obtain \(F_{t_q} \in \mathbb{R}^{L_q \times C}\), while \(\mathcal{A}\) is tokenized into the id sequence \(T_{\mathcal{A}} \in \mathbb{R}^{L_a}\), which serves as the supervision target.

We then introduce two sets of task-specific learnable queries, metric depth queries \(Q_m \in \mathbb{R}^{K \times C}\) and 3D structural queries \(Q_s \in \mathbb{R}^{K \times C}\), each consisting of \(K\) vectors dedicated to capturing metric-depth cues and 3D structural relations, respectively. The two query groups are placed after the visual tokens to form the full LLM input sequence \([F_v,\; Q_m,\; Q_s,\; F_{t_q}]\), so that both query groups can causally attend to the shared visual context. Whether \(F_{t_q}\) appears before or after the queries does not affect the design; we place it after the queries solely for notational simplicity.
Since both query groups participate in the same causal self-attention at every transformer layer while pursuing heterogeneous learning objectives, unconstrained attention from \(Q_s\) to \(Q_m\) would contaminate the structural pathway with metric-depth signals. To prevent such leakage, we introduce a task-decoupled attention mask, as illustrated in Figure~\ref{fig3}. Let \([S_m, E_m]\) and \([S_s, E_s]\) denote the index ranges of the metric and structural queries in the input sequence. We modify the standard causal mask \(\mathbf{M}_{\text{causal}}\) by setting:
\[
    \mathbf{M}_{\text{causal}}[i, j] = -\infty, \quad \forall\, i \in [S_s, E_s],\; j \in [S_m, E_m]
\]
Note that causal masking inherently prevents \(Q_m\) from attending to \(Q_s\); the above constraint further blocks the reverse direction, so that the two query streams are fully decoupled from each other while both still fully attend to the shared visual embeddings, thereby routing metric-depth and 3D structural cues into two independent pathways. After forward propagation through the LLM, the last hidden states at the query positions are extracted as \(\hat{Q}_m, \hat{Q}_s \in \mathbb{R}^{K \times C}\), and the last hidden states at the visual token positions of the \(i\)-th frame are extracted as the LLM-refined patch embeddings \(\hat{f}_v^i \in \mathbb{R}^{P \times C}\). All of them are forwarded to the Expert-Guided Visual Grounding (EVG) module.

\subsection{Expert-Guided Visual Grounding}

Although \(\hat{Q}_m\) and \(\hat{Q}_s\) carry task-relevant context aggregated by the LLM, there is no guarantee that they encode the specific spatial information required for structural or metric perception, so explicit grounding in expert visual knowledge is necessary to activate these capabilities. Moreover, the LLM's abstract semantic space is fundamentally misaligned with the geometric feature spaces demanded by precise spatial perception. To bridge this gap, the Expert-Guided Visual Grounding (EVG) module anchors each query output in frame-level visual features and supervises it with representations from a corresponding pretrained vision foundation model. Since a single linear projection is insufficient for such cross-domain alignment, we adopt a Perceiver-style attention mechanism that first fuses the query output with patch features and then distills task-relevant spatial information through a set of learnable visual queries.

We describe the metric pathway as representative; the structural pathway is processed symmetrically. Given the LLM-refined patch embeddings of the \(i\)-th frame \(\hat{f}_v^i \in \mathbb{R}^{P \times C}\) and the LLM-refined metric queries \(\hat{Q}_m\), we first obtain a query-conditioned visual feature via a linear projection:
\begin{equation}
    f^i_m = \mathrm{Linear}\!\left([\hat{f}_v^i;\hat{Q}_m]\right),
\end{equation}
where \([\cdot\,;\cdot]\) denotes concatenation along the token dimension. We then apply a perceiver attention block \cite{jaegle2021perceiver} followed by an output projection:
\begin{equation}
    f^i_{vq} = \mathrm{Linear}(\mathrm{PerceiverAttn}\left(Q_v, [f^i_m;Q_v]\right)),
\end{equation}
where \(Q_v \in \mathbb{R}^{K_v \times C}\) is a set of learnable visual queries.
The aligned representation $f^i_{vq}$ is then supervised against the expert feature $f^i_{\mathrm{gt}}$ extracted by a pretrained vision foundation model.
We combine an MSE regression loss for point-wise proximity and an InfoNCE contrastive loss \cite{rusak2024infonce} for discriminative alignment:
\begin{equation}
    \mathcal{L}_{\mathrm{MSE}} = \frac{1}{N} \sum_{i=1}^{N}(\left\| f^i_{vq} - f^i_{\mathrm{gt}} \right\|_2^2)
\end{equation}
\begin{equation}
\mathcal{L}_{\mathrm{CL}} = -\frac{1}{N}\sum_{i=1}^{N} \log \frac{\exp\left(\mathrm{sim}(f^i_{vq}, f^i_{\mathrm{gt}})/\tau\right)}{\displaystyle\sum_{f_{\mathrm{gt}} \in \mathcal{B}} \exp\left(\mathrm{sim}(f^i_{vq}, f_{\mathrm{gt}})/\tau\right)},
\end{equation}
where \(\mathcal{B}\) denotes the set of expert features from all frames within the current training batch, \(\mathrm{sim}(\cdot,\cdot)\) is cosine similarity on the flattened representations, and \(\tau\) is a learnable temperature.
The total alignment loss is:
\begin{equation}
    \mathcal{L}_{\mathrm{Align}} = \mathcal{L}_{\mathrm{MSE}} + \lambda \mathcal{L}_{\mathrm{CL}}
\end{equation}
where $\lambda$ balances the two objectives, and we set it to 0.01 based on magnitude considerations. The same procedure is applied symmetrically to $\hat{Q}_s$ under supervision from the structural expert.

\subsection{Two-Stage Training Strategy}

We adopt a two-stage training strategy to progressively equip GAMSI with spatial understanding capabilities. Stage 1 instills foundational 3D structural and metric-depth perception, while Stage 2 generalizes these priors to diverse downstream spatial intelligence tasks. Both stages share an identical optimization objective, which we formalize at the end of this subsection for clarity.

\textbf{Stage 1: Spatial Perception Alignment.}
In the first stage, we train GAMSI on the open-source SenseNova-SI-800K dataset \cite{cai2025scaling} to instill foundational 3D structural and metric-depth perception capabilities. Through the visual alignment supervision imposed by the EVG module, this stage establishes the geometric priors that are essential for subsequent task-conditioned intelligence.

\begin{figure}[t]
\centering
\includegraphics[width=1.0\linewidth]{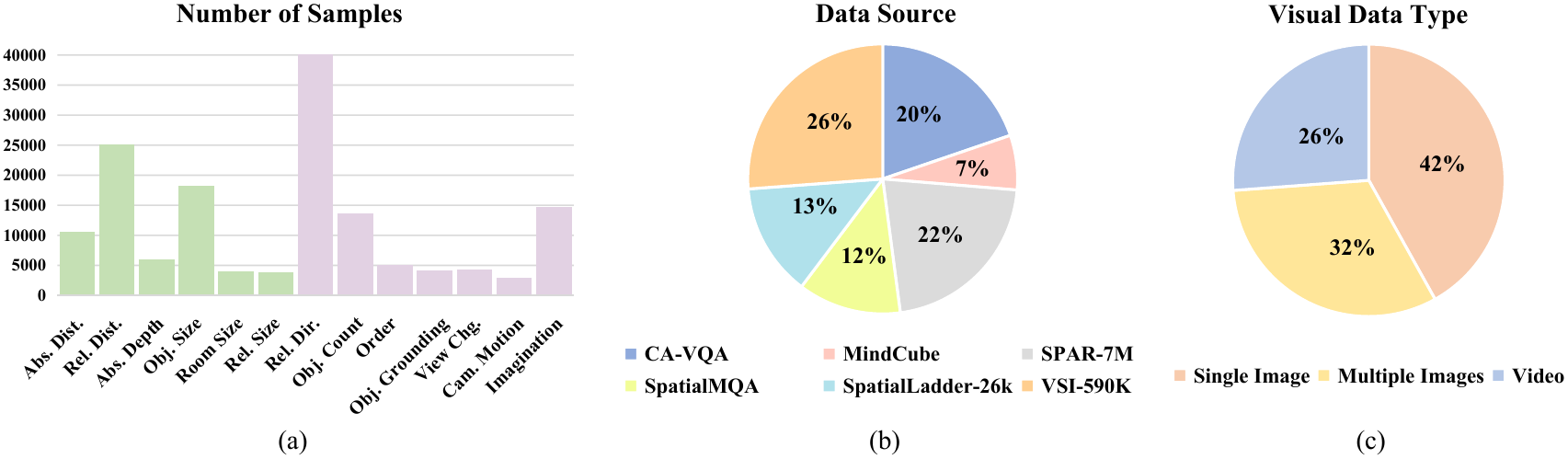}
\caption{Composition of the MTS Dataset. (a) Distribution of Samples across Spatial Task Types; (b) Proportion of samples contributed by each source dataset; (c) Proportion of samples across three visual modalities (single image, multiple images, and video).}
\label{fig:dataset_overview}
\end{figure}

\textbf{Stage 2: Multi-task Spatial Understanding Fine-tuning.}
Building upon the foundational 3D structural and metric-depth perception acquired in Stage 1, the second stage aims to generalize GAMSI's capabilities across a broad spectrum of downstream spatial intelligence tasks. To this end, we construct a large-scale multi-task instruction-tuning corpus, termed the MTS (Multi-Task Spatial) Dataset, by consolidating and re-organizing samples from six publicly available spatial intelligence datasets: CA-VQA \cite{daxberger2025mm}, MindCube \cite{yin2025spatial}, SPAR-7M \cite{zhang2025flatland}, SpatialMQA \cite{liu2025can}, SpatialLadder-26k \cite{li2025spatialladder}, and VSI-590K \cite{yang2025cambrian}.
As summarized in Figure~\ref{fig:dataset_overview}, the MTS Dataset contains $152{,}776$ instruction-following samples covering $13$ spatial task types, and spans three visual modalities in comparable proportions: single image ($42\%$), multiple images ($32\%$), and video ($26\%$). 
To further structure the task space, we organize the $13$ task types into two complementary groups: (i) \emph{Metric Perception} (absolute/relative distance, absolute depth, object- and room-level absolute size, relative size); (ii) \emph{Structural Spatial Perception} (relative direction, object count, appearance order, object grounding, view change inference, camera motion inference, spatial imagination). Together, these two groups cover the spectrum from low-level geometric measurement to high-level structural and viewpoint-aware perception. The MTS Dataset will be publicly released to facilitate future research on unified spatial intelligence.

\textbf{Unified Training Objective.}
Both training stages share an identical optimization objective, which couples the visual alignment supervision from the EVG module with the standard autoregressive language modeling loss over the answer tokens. The language modeling loss maximizes the likelihood of each answer token conditioned on all preceding context:
\begin{equation}
    \mathcal{L}_{\mathrm{LM}} = -\sum_{t=1}^{L_a} \log p\left(T_{\mathcal{A},t} \,\big|\, T_{\mathcal{A},<t},\, F_v,\, Q_s,\, Q_m,\, F_{t_q}\right),
\end{equation}
where $L_a$ is the answer length, $T_{\mathcal{A},t}$ denotes the $t$-th answer token, and $T_{\mathcal{A},<t}$ represents all preceding answer tokens. The overall training objective used in both stages is:
\begin{equation}
    \mathcal{L} = \mathcal{L}_{\mathrm{LM}} + \mathcal{L}_{\mathrm{Align}},
\end{equation}
where $\mathcal{L}_{\mathrm{Align}}$ is the visual alignment loss defined in the EVG. The two stages differ only in the training dataset employed: SenseNova-SI for Stage 1 and the MTS Dataset for Stage 2. The loss formulation and optimization pipeline remain consistent across stages, enabling a smooth transfer of geometric priors from spatial perception learning to multi-task fine-tuning.

\section{Experiments}

\begin{table}[]
\centering
\caption{
Comparison with state-of-the-art methods on seven spatial intelligence benchmarks. All benchmark scores and the macro-average are reported in percentage (\%). ``SI Dataset'' denotes the size of the spatial-intelligence training set used by each method, ``Avg.'' is the macro-average across all benchmarks, and MindCube$^{*}$ refers to the MindCube-Tiny. \textbf{Bold} and \underline{underline} indicate the best and second-best result per column. Subscripts $S\textit{1}$ and $S\textit{2}$ denote Stage-1 and Stage-2, respectively.
}
\label{tab:main_results}
\renewcommand\arraystretch{1.1}
\setlength{\tabcolsep}{1.2pt}
\resizebox{\textwidth}{!}
{
\begin{tabular}{cccccccccc}
\hline
\textbf{Method}           & \textbf{SI Dataset} & \textbf{Avg.} & \textbf{VSI-Bench} & \textbf{MindCube$^*$} & \textbf{ViewSpatial} & \textbf{CV-Bench} & \textbf{SPBench-SI} & \textbf{SPBench-MV} & \textbf{SparBench} \\ \hline
\hline 
\multicolumn{10}{l}{\cellcolor[HTML]{EFEFEF}\textit{\textbf{Open-source General Models}}}                                                                                                                                 \\ \hline
\hline 
Qwen2.5-VL-3B-Instruct \cite{bai2025qwen3}    & -                   & 40.3          & 27.5               & 42.5                   & 36.2                 & 63.0              & 40.9                & 41.8                & 30.0               \\
Bagel-7B-MoT \cite{deng2025emerging}              & -                   & 40.7          & 32.5               & 34.2                   & 41.3                 & 56.7              & 38.1                & 43.1                & 39.1               \\
InternVL3-2B \cite{zhu2025internvl3}             & -                   & 40.9          & 33.5               & 36.8                   & 32.7                 & 68.1              & 41.2                & 46.7                & 27.2               \\
Qwen2.5-VL-7B-Instruct \cite{bai2025qwen3}    & -                   & 43.4          & 33.3               & 34.2                   & 37.4                 & 75.3              & 47.4                & 42.7                & 33.8               \\
Qwen3-VL-2B-Instruct \cite{bai2025qwen3}      & -                   & 46.8          & 48.5               & 35.7                   & 36.4                 & 66.3              & 51.6                & 55.9                & 33.3               \\
InternVL3-8B \cite{zhu2025internvl3}             & -                   & 47.5          & 43.3               & 41.8                   & 38.6                 & 68.4              & 51.5                & 52.9                & 35.8               \\ 
Qwen3-VL-8B-Instruct \cite{bai2025qwen3}      & -                   & 52.7          & 56.9               & 29.9                   & 42.2                 & 82.4              & 51.7                & 65.7                & 39.9               \\
\hline
\hline 
\multicolumn{10}{l}{\cellcolor[HTML]{EFEFEF}\textit{\textbf{Open-source Spatial Intelligence Models}}}                                                                                                                    \\ \hline
\hline 
SpatialLadder-3B \cite{li2025spatialladder}          & 260K                & 54.5          & 45.1               & 43.5                   & 44.8                 & 73.6              & {\ul 70.0}                & 71.7                & 32.9               \\
Spatial-MLLM-4B \cite{wu2025spatial}           & 135K                & 43.8          & 46.2               & 33.5                   & 34.6                 & 54.7              & 45.3                & 56.7                & 35.3               \\
SpaceR-7B \cite{ouyang2025spacer}                 & 151K                & 46.8          & 40.9               & 32.1                   & 38.4                 & 74.9              & 47.9                & 56.2                & 37.3               \\
ViLaSR-7B \cite{wu2025reinforcing}                 & 81K                 & 43.3          & 44.9               & 35.2                   & 35.7                 & 43.5              & 48.9                & 57.5                & 37.3               \\
VST-3B-SFT \cite{yang2025visual}               & 4M                  & 54.6          & 51.7               & 36.0                   & 52.9                 & 84.6              & 53.7                & 65.0                & 38.1               \\
VST-7B-SFT \cite{yang2025visual}                & 4M                  & 57.0          & 55.5               & 39.1                   & 50.7                 & {\ul 85.5}              & 52.2                & 68.5                & 47.2               \\
Cambrian-S-3B \cite{yang2025cambrian}             & 590K                & 51.0          & 55.9               & 39.0                   & 40.9                 & 75.2              & 57.7                & 55.4                & 32.9               \\
Cambrian-S-7B \cite{yang2025cambrian}            & 590K                & 52.3          & 63.0               & 38.1                   & 41.3                 & 76.9              & 54.6                & 54.4                & 38.0               \\
SenseNova-SI-Qwen3-VL-8B \cite{cai2025scaling}  & 8M                  & 63.4          & 62.9               & 73.8                   & 51.2                 & 83.4              & 60.4                & 71.1                & 41.1               \\
SenseNova-SI-InternVL3-8B \cite{cai2025scaling} & 800K                & 64.0          & 60.6               & 66.7                   & 55.1                 &84.0        & 62.5          & 70.3                & 48.9               \\ \hline
\hline 
\multicolumn{10}{l}{\cellcolor[HTML]{EFEFEF}\textit{\textbf{Ours}}}                                                                                                                    \\ \hline
\hline 
         GAMSI$_{S\textit{1}}$                 & 800K                & {\ul 67.4}    & {\ul 66.0}         & {\ul 75.1}             & {\ul 59.6}           & 83.9              & 60.6                & {\ul 76.9}          & {\ul 49.6}         \\
        GAMSI$_{S\textit{1+}S\textit{2}}$                  & 950K                & \textbf{75.8} & \textbf{68.5}      & \textbf{94.1}          & \textbf{63.0}        & \textbf{85.7}     & \textbf{79.1}       & \textbf{80.4}       & \textbf{59.9}      \\ \hline
\end{tabular}}
\end{table}

\subsection{Experimental Settings}
\textbf{Implementation Details.} We build all our experiments upon Qwen3-VL-8B-Instruct \cite{bai2025qwen3}, and adopt a two-stage supervised fine-tuning pipeline. We set the number of metric and structural queries to $K=40$ each. Stage 1 trains on SenseNova-SI-800K \cite{cai2025scaling} to jointly aligns the multimodal architecture and injects general spatial-intelligence priors into the model; this stage is optimized for up to 3 epochs with a batch size of 64 and a learning rate of \(8 \times 10^{-6}\). In the second stage, we curate the MTS Dataset by combining CA-VQA \cite{daxberger2025mm}, MindCube \cite{yin2025spatial}, SPAR-7M \cite{zhang2025flatland}, SpatialMQA \cite{liu2025can}, SpatialLadder-26K \cite{li2025spatialladder}, and VSI-590K \cite{yang2025cambrian}, and continue fine-tuning for 2 epochs with a batch size of 64 on 8 NVIDIA H200 GPUs and a learning rate of \(4 \times 10^{-6}\). For both stages, we employ the AdamW optimizer.

\textbf{Evaluation Benchmarks.} We evaluate GAMSI on seven widely-used spatial-intelligence benchmarks using the EASI toolkit \cite{cai2025holistic}, namely VSI-Bench \cite{yang2025thinking} (5,155 video questions), MindCube-Tiny \cite{yin2025spatial} (1,040 multi-image questions on viewpoint transformation and spatial imagination), ViewSpatial-Bench \cite{li2025viewspatial} (5,712 questions for perspective-dependent reasoning), CV-Bench \cite{tong2024cambrian} (2,638 single-image questions on 2D/3D vision tasks), SPBench-SI/SPBench-MV \cite{li2025spatialladder} (1,009 single-image and 319 multi-image questions), and SPAR-Bench \cite{zhang2025flatland} (7,211 questions for multi-difficulty spatial reasoning).

\subsection{Main Results}

Table~\ref{tab:main_results} compares GAMSI with open-source general-purpose MLLMs \cite{bai2025qwen3, deng2025emerging,zhu2025internvl3} and spatial-intelligence-specialized models \cite{li2025spatialladder,wu2025spatial,ouyang2025spacer,wu2025reinforcing,yang2025visual,yang2025cambrian,cai2025scaling} on seven benchmarks covering single-image, multi-image, and video inputs. Overall, GAMSI$_{S\textit{1+}S\textit{2}}$ achieves the best result on every benchmark and improves the macro-average from $64.0\%$ (the strongest prior model, SenseNova-SI-InternVL3-8B) to $75.8\%$, an absolute improvement of $11.8\%$. The largest margins over the best prior result on each benchmark are observed on benchmarks dominated by relational and viewpoint-dependent reasoning ($+20.3\%$ on MindCube-Tiny over SenseNova-SI-Qwen3-VL-8B, $+11.0\%$ on SPAR-Bench over SenseNova-SI-InternVL3-8B, and $+9.1\%$ on SPBench-SI over SpatialLadder-3B), corresponding to the regime in which 2D semantic features alone are insufficient and in which the proposed 3D structural pathway is most effective. The advantage of GAMSI is further corroborated from two complementary, partially controlled angles. Under the same SenseNova-SI-800K training data, GAMSI$_{S\textit{1}}$ outperforms SenseNova-SI-InternVL3-8B on most benchmarks and improves the macro-average from $64.0\%$ to $67.4\%$ ($+3.4\%$), demonstrating that, when the data scale is held fixed, the proposed metric-structure decoupled pathways and EVG grounding yield gains that purely data-driven instruction tuning cannot achieve. Under the identical Qwen3-VL-8B backbone, GAMSI$_{S\textit{1}}$ further surpasses SenseNova-SI-Qwen3-VL-8B by $4.0\%$ on the macro-average ($67.4\%$ vs.\ $63.4\%$) while consuming roughly $10\times$ less spatial data ($800$K vs.\ $8$M), with the largest gains again on structure- and viewpoint-heavy tasks ($+8.4\%$ on ViewSpatial, $+8.5\%$ on SPAR-Bench, $+5.8\%$ on SPBench-MV, and $+3.1\%$ on VSI-Bench), indicating that injecting an appropriate architectural prior is a substantially more sample-efficient route to spatial competence than scaling training data alone. Building on this foundation, Stage-2 multi-task fine-tuning on the MTS Dataset further improves the macro-average from $67.4\%$ to $75.8\%$ ($+8.4\%$), with the largest gains concentrated on benchmarks well aligned with our $13$ task types ($+19.0\%$ on MindCube-Tiny, $+18.5\%$ on SPBench-SI, and $+10.3\%$ on SPAR-Bench), confirming that the foundational metric and structural priors instilled in Stage-1 are not benchmark-specific and transfer effectively across task formulations and input modalities once exposed to a diverse, well-organized multi-task instruction dataset.

\subsection{Ablation Experiment}

To dissect the contribution of each component in GAMSI, we conduct controlled ablations on a balanced 10K subset sampled from SenseNova-SI-800K. All variants share the Qwen3-VL-8B backbone and identical training hyper-parameters, and only the components introduced by MSDQ and EVG are toggled. Results on the seven spatial-intelligence benchmarks are reported in Table~\ref{tab:ablation}.

\begin{table}[t]
\centering
\caption{
Ablation study of the dual-pathway design on a 10K balanced subset of SenseNova-SI-800K. All benchmark scores and the macro-average are reported in percentage (\%). ``$Q_s$'' and ``$Q_m$'' denote the 3D structural and metric-depth query pathways. ``Mask'' indicates the task-decoupled attention mask in MSDQ. MindCube$^{*}$ refers to the MindCube-Tiny subset. \textbf{Bold} marks the best result.
}
\label{tab:ablation}
\renewcommand\arraystretch{1.2}
\setlength{\tabcolsep}{1.2pt}
\resizebox{\textwidth}{!}
{
\begin{tabular}{ccccccccccc}
\hline
$Q_s$ & $Q_m$ & Mask & \textbf{Avg.} & \textbf{VSI-Bench} & \textbf{MindCube$^*$} & \textbf{ViewSpatial} & \textbf{CV-Bench} & \textbf{SPBench-SI} & \textbf{SPBench-MV} & \textbf{SparBench} \\ \hline
-    & -    & -          & 54.1          & 55.0          & 28.6          & 42.3          & 86.3 & 50.8          & 68.9          & {46.9} \\
$\checkmark$   & - & -         & 55.0          & 56.4          & {35.7} & {43.8} & 81.6          & 52.3          & 68.6          & 46.5          \\
$\checkmark$ & $\checkmark$ & -  & 55.1          & {56.9} & 33.8          & 43.0          & 83.2          & {52.8} & 68.9          & 46.8          \\
$\checkmark$ & $\checkmark$ & $\checkmark$ & \textbf{55.7} & 56.4         & {35.7} & 42.1          & {86.3} & 52.3          & {69.9} & {46.9} \\ \hline
\end{tabular}}
\end{table}

\textbf{Effect of dual-pathway geometric priors.} Adding the 3D structural pathway alone improves the macro-average from $54.1\%$ to $55.0\%$ ($+0.9\%$), with the largest gains concentrated on benchmarks that explicitly target viewpoint transformation and perspective-dependent reasoning ($+7.1\%$ on MindCube-Tiny and $+1.5\%$ on ViewSpatial), together with a moderate gain on VSI-Bench ($+1.4\%$). This pattern confirms that VGGT-grounded structural cues address the regime in which 2D semantic features alone are insufficient. However, this single-pathway variant regresses on CV-Bench (from $86.3\%$ to $81.6\%$), a benchmark in which 3D depth and 3D distance subtasks account for a substantial portion of the questions, indicating that structural priors alone are inadequate, and may even be detrimental, for absolute scale perception. Activating the metric-depth pathway under task-decoupled mask isolation on top further improves the macro-average to $55.7\%$ and restores CV-Bench to $86.3\%$ ($+4.7\%$ over the structural-only variant), while largely preserving the structural gains, with only a marginal drop on ViewSpatial. This confirms that structural and metric priors address disjoint subsets of spatial questions, and their joint internalization yields balanced spatial cognition.

\textbf{Effect of the task-decoupled attention mask.} Comparing the two dual-pathway variants, removing the mask reduces the macro-average from $55.7\%$ to $55.1\%$ ($-0.6\%$), and produces consistent regressions across most benchmarks ($-3.1\%$ on CV-Bench, $-1.9\%$ on MindCube-Tiny, and $-1.0\%$ on SPBench-MV). This indicates that, under unconstrained causal self-attention, signals from $Q_m$ leak into $Q_s$ and contaminate each pathway with the other's objective, eroding the gains from injecting two heterogeneous priors. The task-decoupled mask separates the two streams at the representation level, restoring the full benefit of dual-pathway routing without introducing additional parameters.

\subsection{Qualitative Visualization}

\begin{figure}[t]
  \centering
  \includegraphics[width=0.74\textwidth]{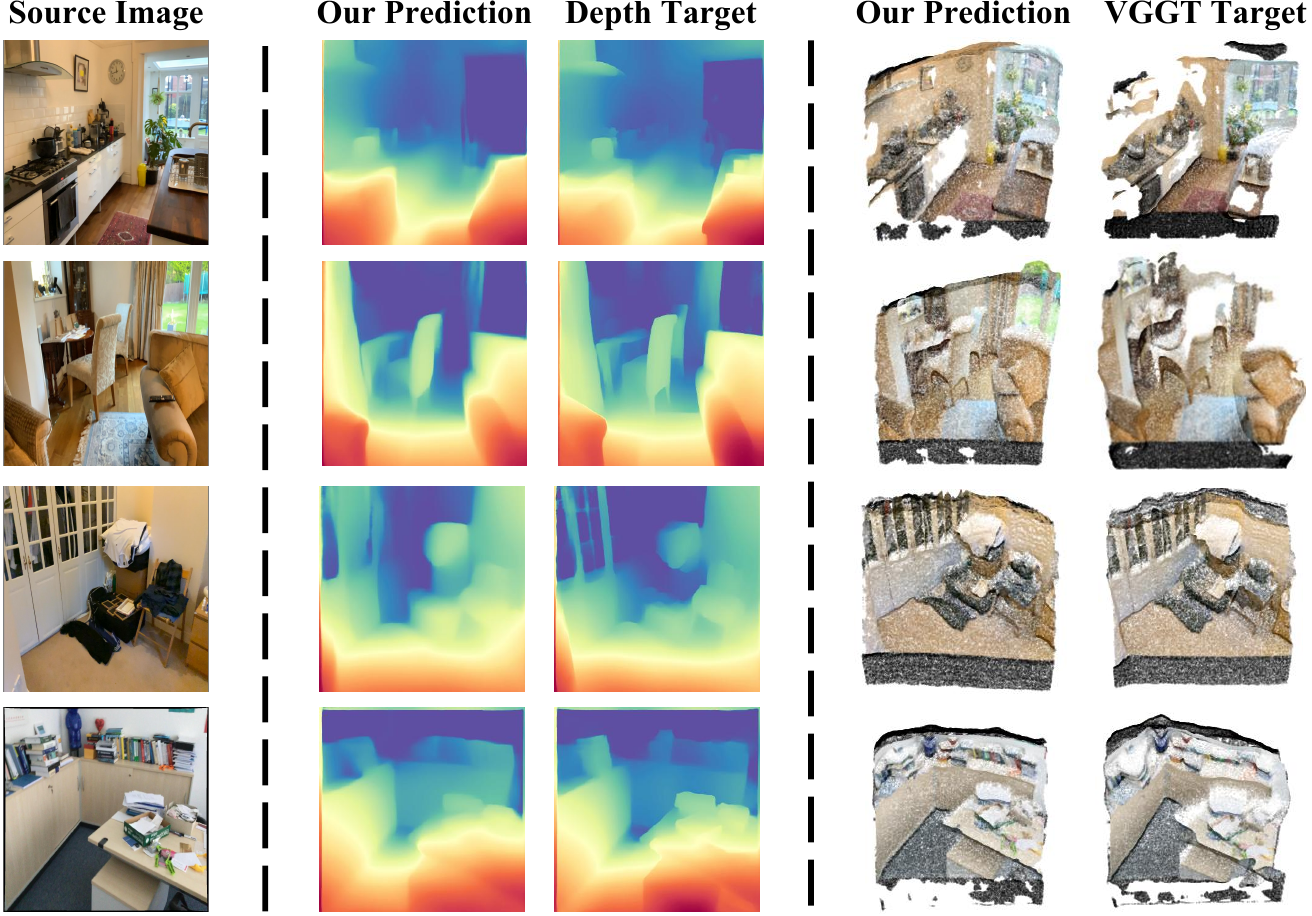}
  \caption{Qualitative comparison of the dual-pathway predictions ($Q_m$ depth and $Q_s$ point cloud) against their respective depth and VGGT targets on four scenes.}
  \label{fig4}
\end{figure}

To verify that the two pathways indeed extract the geometric information they are designed for, we visualize the predictions of $Q_m$ and $Q_s$ on four indoor scenes in Figure~\ref{fig4}, where the \emph{Depth Target} and \emph{VGGT Target} are produced by the EVG experts as supervision. The depth maps decoded from $Q_m$ closely match the depth target: distant regions such as windows are assigned cool tones while nearby furniture is rendered in warm tones, with no noticeable blurring or scale drift and only minor, barely perceptible differences from the target. In parallel, the point clouds decoded from $Q_s$ faithfully reproduce both the global scene layout and the camera viewpoint of the VGGT target, including furniture arrangement, wall and floor orientation, and finer details such as plants and books. Moreover, the two pathways behave as complementary rather than redundant channels: $Q_m$ does not produce sparse structural artifacts, and $Q_s$ does not encode dense pixel-wise depth, indicating that the task-decoupled attention mask successfully isolates the two streams. Together, these observations confirm that $Q_m$ carries metric depth and $Q_s$ carries 3D structural information, validating the dual-pathway design at the representation level.

\section{Conclusion}

We presented GAMSI, a dual-pathway geometry-aware MLLM that internalizes two complementary forms of geometric knowledge, namely fine-grained metric depth and holistic 3D structural perception, into a unified autoregressive backbone, while taking only RGB images as input at inference time. The Metric-Structure Decoupled Queries (MSDQ) employ two groups of learnable queries together with a task-decoupled attention mask to cleanly disentangle the metric and structural pathways at the representation level, and the Expert-Guided Visual Grounding (EVG) module aligns the queried cues with vision foundation models as training-time-only supervision, eliminating any inference-time dependence on auxiliary inputs. To foster multi-task tuning for higher-order spatial intelligence, we further construct MTS, a $152{,}776$-sample dataset covering 13 task types and three visual modalities. Trained with a two-stage curriculum, GAMSI achieves state-of-the-art results on seven spatial intelligence benchmarks, with ablations and visualizations confirming that the two pathways capture their intended metric and structural information.

\section{Limitations}
\label{limitations}

Despite the promising results, the proposed GAMSI has several limitations: (1) While GAMSI performs strongly on a wide range of spatial intelligence tasks, its behavior in more complex scenarios (e.g., long-horizon spatial planning), which jointly demand fine-grained geometric perception and high-level reasoning, warrants deeper investigation. (2) Constrained by computational resources, our experiments are conducted on an 8B-scale backbone; the scalability of GAMSI to larger models and its adaptability to diverse multimodal architectures remain to be explored.

\bibliography{references}
\bibliographystyle{unsrt}

\newpage

\appendix

\section{Details of the MTS Dataset}
\label{app:mts_dataset}

To support comprehensive spatial reasoning capabilities in our model, we construct the Multi-Task Spatial (MTS) instruction tuning dataset by consolidating and re-organizing samples from six publicly available spatial intelligence datasets, namely CA-VQA \cite{daxberger2025mm}, MindCube \cite{yin2025spatial}, SPAR-7M \cite{zhang2025flatland}, SpatialMQA \cite{liu2025can}, SpatialLadder-26K \cite{li2025spatialladder}, and VSI-590K \cite{yang2025cambrian}. The resulting corpus contains 152{,}776 instruction-following samples spanning 13 fine-grained spatial task types. Table~\ref{tab:mts_stats} reports the detailed composition of MTS, including the source dataset(s) and the number of samples associated with each task type.

\subsection{Construction Pipeline}

Since each source dataset is already a well-curated spatial intelligence training dataset, we only perform the following lightweight pre-processing steps on each of them to obtain a unified instruction-following format:

\textbf{Task mapping.} We manually inspect the original category annotations of each source dataset. For datasets without explicit category labels, we perform keyword-based matching according to the characteristics of the questions to assign a category. We then select samples belonging to our 13 target task types from the categorized pool.

\textbf{Format unification.} All samples are converted into a consistent dictionary format consisting of the system-user dialogue (\texttt{messages}), the task type (\texttt{type}), and the associated \texttt{images}.

\begin{table}[t]
\centering
\caption{Statistics of the MTS Dataset by task type. The dataset contains 152{,}776 instruction-following samples covering 13 spatial task types consolidated from six public spatial intelligence datasets.}
\label{tab:mts_stats}
\renewcommand\arraystretch{1.2}
\resizebox{\linewidth}{!}{
\begin{tabular}{c|c|c}
\hline
\textbf{Type}        & \textbf{Data Source}                                   & \textbf{Number} \\ \hline
absolute distance & CA-VQA, SPAR-7M, SpatialLadder-26k & 10{,}629 \\
relative distance & CA-VQA, SPAR-7M, SpatialLadder-26k, VSI-590K & 25{,}109 \\
absolute depth & SPAR-7M & 6{,}000 \\
absolute size (object) & CA-VQA, SpatialLadder-26k, VSI-590K & 18{,}229 \\
absolute size (room) & SpatialLadder-26k, VSI-590K & 3{,}959 \\
relative size & CA-VQA & 3{,}847 \\
relative direction & CA-VQA, SPAR-7M, SpatialMQA, SpatialLadder-26k, VSI-590K & 40{,}102\\
object count & CA-VQA, SpatialLadder-26k, VSI-590K & 13{,}664\\
appearance order & SpatialLadder-26k, VSI-590K & 5{,}126\\
object grounding & CA-VQA, SPAR-7M & 4{,}111\\
view change inference & MindCube, SPAR-7M & 4{,}302\\
camera motion inference & SPAR-7M & 3{,}000\\
spatial imagination & MindCube, SPAR-7M & 14{,}698\\ \hline
13 Types             & CA-VQA, MindCube, SPAR-7M, SpatialMQA, SpatialLadder-26k, VSI-590K  & 152{,}776          \\ \hline
\end{tabular}}
\end{table}

\subsection{Data Source Summary}

The six source datasets contribute complementary aspects to MTS:

\textbf{CA-VQA} provides a broad mixture of metric and qualitative spatial questions, contributing to 7 of the 13 task types.

\textbf{SPAR-7M} offers large-scale pixel-level and viewpoint-level annotations, covering 8 task types including depth, grounding, view change, camera motion, and spatial imagination.

\textbf{SpatialLadder-26k} supplies hierarchical spatial questions ranging from object counting to absolute distance estimation.

\textbf{VSI-590K} focuses on video-based spatial intelligence and significantly enlarges the pool of relative direction, object count, and room-level size samples.

\textbf{SpatialMQA} contributes relative direction questions.

\textbf{MindCube} provides mental rotation and viewpoint-transformation samples that are crucial for the spatial imagination and view change inference tasks.

\section{Detailed Experimental Settings}

\subsection{Details of Visual Expert model}

In our implementation, we instantiate the metric and structural experts with Depth Anything V2~\cite{yang2024depthv2} and VGGT~\cite{wang2025vggt}, respectively.

\subsection{Details of Evaluation Benchmarks}

\textbf{VSI-Bench.} \cite{yang2025thinking} is a comprehensive evaluation benchmark designed to assess visual-spatial intelligence in Multimodal Large Language Models (MLLMs) through egocentric video understanding. The benchmark comprises 5,155 video question-answer pairs derived from 288 real-world videos. The videos span diverse environments across multiple geographic regions, providing broad coverage for evaluating spatial understanding in realistic scenarios. By leveraging continuous egocentric video streams rather than isolated images, VSI-Bench targets the spatial integration abilities required for embodied scene understanding.

\textbf{MindCube-Tiny} \cite{yin2025spatial} is a multi-image evaluation benchmark comprising 1,040 question-answer pairs focused on viewpoint transformation and spatial imagination. It assesses models' abilities to mentally manipulate spatial configurations across different viewpoints, requiring integrated reasoning over multiple images to infer spatial relationships not directly observable from any single view. MindCube-Tiny is organized into three settings of increasing complexity: (1) Rotation, where a stationary camera rotates in place to capture 2 to 4 orthogonal views of a central foreground object; (2) Occlusion, which leverages occlusion phenomena to require object permanence under partial visibility, transformation of lateral (left-right) relationships into depth (front-back) relationships across views, and multi-view integration for coherent 3D understanding; and (3) Among, where the camera rotates around a central object positioned between it and several surrounding objects, capturing four orthogonal views that each place the central object in front of one surrounding object.

\textbf{ViewSpatial-Bench} \cite{li2025viewspatial} is a comprehensive evaluation framework comprising 5,712 question-answer pairs across more than 1,000 3D scenes. This benchmark evaluates VLMs' spatial localization capabilities from both egocentric and allocentric viewpoints, addressing the critical gap in perspective-dependent reasoning abilities that are essential for embodied interaction and multi-agent collaboration. By jointly probing first-person and third-person perspectives, ViewSpatial-Bench provides a systematic testbed for evaluating whether models can reconcile spatial references across different observer frames.

\textbf{CV-Bench} \cite{tong2024cambrian} addresses the limitations of existing vision-centric benchmarks through 2,638 manually-inspected single-image examples. The benchmark repurposes established vision datasets to evaluate MLLMs on fundamental 2D and 3D computer vision tasks. The evaluation encompasses 2D spatial comprehension through spatial relationships and object counting, while 3D understanding is assessed via depth ordering and relative distance estimation. By repurposing well-curated vision datasets, CV-Bench ensures high annotation reliability and offers a principled reference point for measuring foundational visual-spatial competence.

\textbf{SPBench-SI \& SPBench-MV} \cite{li2025spatialladder} are evaluation benchmarks constructed using the SpatialLadder-26k pipeline applied to the ScanNet validation set. Both benchmarks undergo rigorous quality control through standard pipeline filtering strategies supplemented by manual curation to ensure data disambiguation and high-quality annotations. SPBench-SI serves as a single-image evaluation benchmark designed to assess models' spatial understanding and reasoning capabilities from individual viewpoints, encompassing four task categories—absolute distance, object size, relative distance, and relative direction—with a total of 1,009 samples. SPBench-MV constitutes a multi-image evaluation benchmark that requires models to perform joint spatial modeling across multiple viewpoints, with a total of 319 samples. In addition to the tasks included in SPBench-SI, SPBench-MV incorporates object counting tasks to evaluate models' capabilities in identifying and enumerating objects within multi-view scenarios.

\textbf{SPAR-Bench} \cite{zhang2025flatland} constitutes a comprehensive evaluation framework for systematically assessing spatial perception and reasoning capabilities in Vision-Language Models (VLMs) across multiple difficulty levels. The benchmark encompasses 20 diverse spatial understanding tasks spanning single-view, multi-view, and video-based modalities, incorporating 7,211 manually verified question-answer pairs to ensure annotation quality and reliability. By covering tasks of varying granularity and input modality, SPAR-Bench enables fine-grained diagnosis of model strengths and weaknesses across the full spectrum of spatial reasoning challenges encountered in real-world visual understanding.

\subsection{Ablation on the Number of Queries}
\label{app:k_ablation}

To examine the effect of the query count $K$ in MSDQ, we vary $K \in \{8, 16, 24, 32, 40, 48, 56\}$ for both metric and structural query groups while keeping all other hyper-parameters fixed. Following the protocol of the main ablation (Table~\ref{tab:ablation}), all variants are trained on the same 10K balanced subset of SenseNova-SI-800K. As reported in Table~\ref{tab:k_ablation}, the macro-average rises from $54.3\%$ at $K \leq 16$ to a peak of $55.7\%$ at $K=40$, then plateaus at $55.2\%$ for $K \geq 48$ with no further benefit, indicating that further enlarging $K$ adds computational overhead without measurable gain. We therefore adopt $K=40$ as the default in all main experiments.

\begin{table}[h]
\centering
\caption{Ablation on the number of learnable queries $K$ per group in MSDQ, evaluated on the same 10K balanced subset of SenseNova-SI-800K used in Table~\ref{tab:ablation}. All scores are reported in percentage (\%). \textbf{Bold} marks the best result per column.}
\label{tab:k_ablation}
\renewcommand\arraystretch{1.2}
\setlength{\tabcolsep}{3pt}
\resizebox{\textwidth}{!}{
\begin{tabular}{c|cccccccc}
\hline
$K$ & \textbf{Avg.} & \textbf{VSI-Bench} & \textbf{MindCube$^*$} & \textbf{ViewSpatial} & \textbf{CV-Bench} & \textbf{SPBench-SI} & \textbf{SPBench-MV} & \textbf{SparBench} \\ \hline
8  & 54.3          & 56.0          & 31.6          & 42.2          & 82.5          & 52.1          & 69.4          & 46.3          \\
16 & 54.3          & {56.6} & 32.2          & 43.1          & 82.4          & {52.9} & 65.9          & 46.7          \\
24 & 54.1          & 56.5          & 32.9          & 43.1          & 81.4          & 51.5          & 67.0          & 46.6          \\
32 & 55.0          & 56.0          & 35.2          & 42.6          & 81.8          & 52.3          & {70.2} & 46.8          \\
40 & \textbf{55.7} & 56.4          & 35.7          & 42.1          & {86.3} & 52.3          & 69.9          & 46.9          \\
48 & 55.2          & {56.6} & 35.0          & 43.1          & 83.2          & 52.7          & 68.9          & {47.1} \\
56 & 55.2          & 56.4          & {38.0} & {43.4} & 82.9          & 51.9          & 67.4          & 46.7          \\ \hline
\end{tabular}}
\end{table}

\section{Broader Impacts}
\label{Broader_Impacts}
\textbf{Positive Societal Impacts.} Our work advances the spatial cognition of multimodal large language models, enabling reliable reasoning about 3D structure and metric scale directly from RGB images. By removing the inference-time dependence on auxiliary depth maps, point clouds, or camera parameters, GAMSI lowers the hardware and sensor requirements for deploying spatially intelligent systems, which could democratize access to such capabilities in resource-constrained settings and benefit applications such as assistive technologies for visually impaired users, safer robotic navigation, and AR/VR interaction.

\textbf{Negative Societal Impacts.} Despite these benefits, GAMSI produces quantitative spatial estimates (e.g., distances and sizes) in natural language, and such outputs may appear authoritative even when incorrect, particularly in out-of-distribution scenes; over-reliance in safety-critical pipelines such as autonomous driving or robotic manipulation could lead to harmful decisions. Moreover, stronger spatial cognition from raw RGB inputs lowers the effort required to infer 3D scene layouts from casually captured images or publicly shared videos, raising privacy concerns regarding the potential reconstruction of private environments without consent. To mitigate these risks, we recommend that deployments include calibrated uncertainty estimation and human oversight in high-stakes decision-making.


\end{document}